%% file: main.tex
\documentclass[10pt,twocolumn,letterpaper]{article}

\usepackage[pagenumbers]{cvpr} 

\usepackage{multirow}
\usepackage{amsmath}
\usepackage{graphicx}
\usepackage{makecell}
\usepackage{colortbl}
\usepackage{booktabs}
\usepackage{algorithm}
\usepackage{algpseudocode}
\usepackage{float}
\usepackage{placeins}
\definecolor{cvprblue}{rgb}{0.21,0.49,0.74}
\usepackage[pagebackref,breaklinks,colorlinks,allcolors=cvprblue]{hyperref}

\def\paperID{****} 
\def\confName{CVPR}
\def\confYear{2026}

\title{HSI-VAR: Rethinking Hyperspectral Restoration through Spatial–Spectral Visual Autoregression}

\hypersetup{
    colorlinks=true,
    urlcolor=magenta
}
\author{
Xiangming Wang$^{1}$\hspace{0.3cm}
Benteng Sun$^{1}$\hspace{0.3cm}
Yungeng Liu$^{1}$\hspace{0.3cm}
Haijin Zeng$^{2,*}$\hspace{0.3cm}
Yongyong Chen$^{1,}$\thanks{Corresponding author.}\hspace{0.2cm}\\
Jingyong Su$^{1}$\hspace{0.2cm}
Jie Liu$^{1}$\\
$^{1}$ Harbin Institute of Technology (Shenzhen)\hspace{0.2cm} $^{2}$ Harvard University \\
Code: \url{github.com/xianggkl/HSI-VAR}
}

\begin{document}
\maketitle
\input{sec/0_abstract}    
\input{sec/1_intro}
\input{sec/2_relatedworks}
\input{sec/3_methods}
\input{sec/4_experiments}
\input{sec/5_conclusion}

{
    \small
    \bibliographystyle{ieeenat_fullname}
    \bibliography{main}
}


\end{document}

%% file: sec/0_abstract.tex
\begin{abstract}

{Hyperspectral images (HSIs) capture richer spatial–spectral information beyond RGB, yet real-world HSIs often suffer from a composite mix of degradations, such as noise, blur, and missing bands.
Existing generative approaches for HSI restoration like diffusion models require hundreds of iterative steps, making them computationally impractical for high-dimensional HSIs.
While regression models tend to produce oversmoothed results, failing to preserve critical structural details.
We break this impasse by introducing HSI-VAR, rethinking HSI restoration as an autoregressive generation problem, where spectral and spatial dependencies can be progressively modeled rather than globally reconstructed.
HSI-VAR incorporates three key innovations: (1) Latent–condition alignment, which couples semantic consistency between latent priors and conditional embeddings for precise reconstruction; (2) Degradation-aware guidance, which uniquely encodes mixed degradations as linear combinations in the embedding space for automatic control, remarkably achieving a nearly $50\%$ reduction in computational cost at inference; (3) A spatial–spectral adaptation module that refines details across both domains in the decoding phase.
Extensive experiments on nine all-in-one HSI restoration benchmarks confirm HSI-VAR's state-of-the-art performance, achieving a 3.77 dB PSNR improvement on \textbf{\textit{ICVL}} and offering superior structure preservation with an inference speed-up of up to $95.5 \times$ compared with diffusion-based methods, making it a highly practical solution for real-world HSI restoration.
Codes are provided in the supplementary materials.}


\end{abstract}

%% file: sec/1_intro.tex
\section{Introduction}
\label{sec:intro}

Hyperspectral Images (HSIs), which inherently consist of significantly more bands in the spectral dimension than standard RGB images, naturally yield richer spectral information and a more comprehensive visual perspective \cite{fu2024ssumamba,lai2023hybrid,wang2017hyperspectral,miao2022hyperspectral,wang2025otlrm,zeng2024unmixing, pan2024diffsci, 10713101}. 
Consequently, they possess important and widespread applications across fields such as autonomous driving \cite{basterretxea2021hsi}, unmanned aerial vehicle inspection \cite{adao2017hyperspectral, lu2020recent}, agricultural production \cite{weber2018hyperspectral} and environmental monitoring \cite{stuart2019hyperspectral}.
Beyond interferences during the acquisition and transmission processes such as noise and blurring, HSIs are also susceptible to spectral-specific confounding factors, such as the missing of spectral bands.
Consequently, this multi-degradation situation severely degrades the quality of the HSIs, constraining subsequent high-level image processing tasks for HSI understanding and generation \cite{wang2025hsi, pang2024hsigene, wang2025hypersigma}.

\begin{figure}
    \centering
    \hspace{3mm}
    \includegraphics[width=0.97\linewidth]{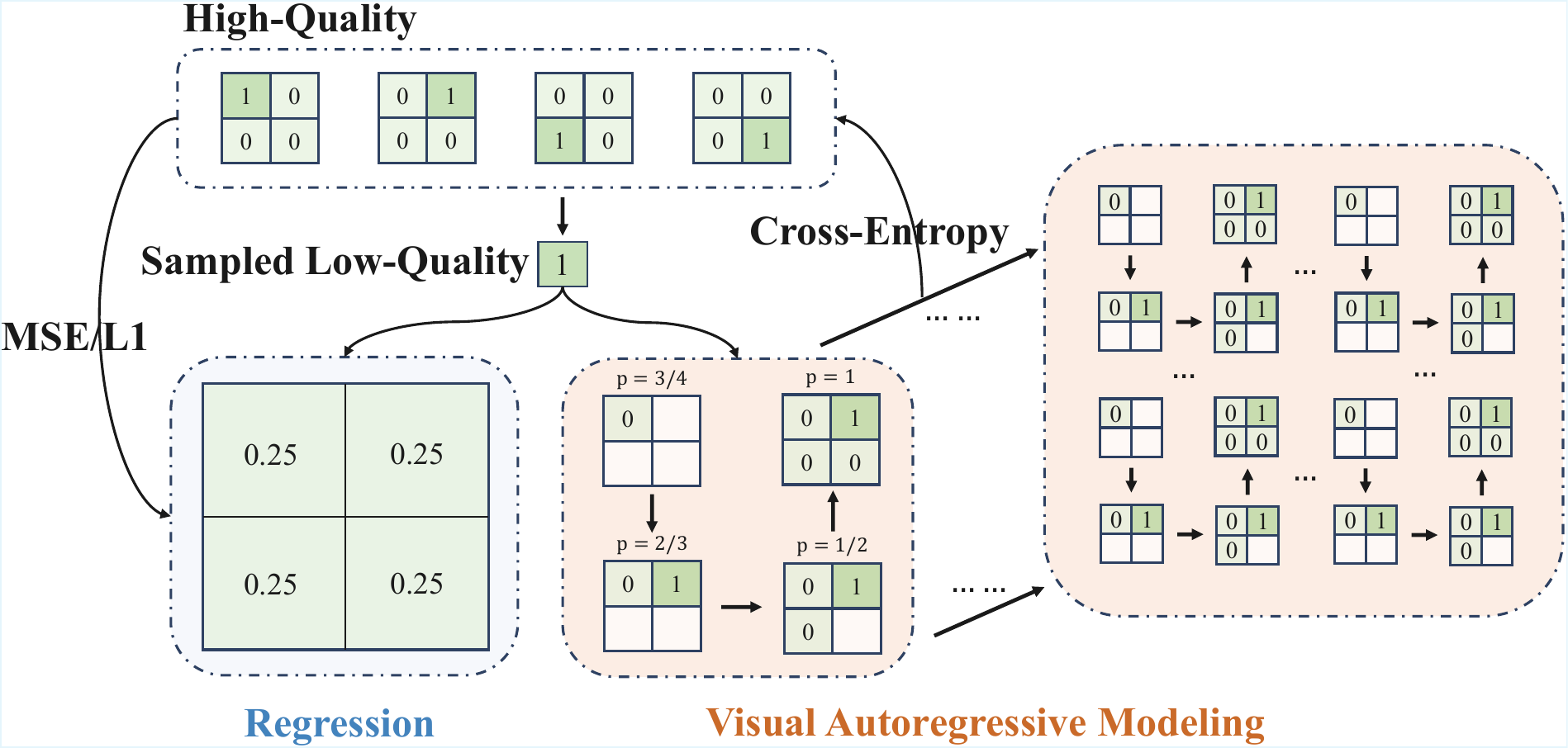}
    \vspace{-2mm}
    \caption{Regression v.s. Visual Autoregressive Modeling. Our HSI-VAR rethinks HSI restoration as scale-wise autoregressive generation, progressively modeling the structure dependencies.}
    \vspace{-6mm}
    \label{fig:highlight}
\end{figure}


Existing deep learning-based all-in-one image restoration methods can be broadly categorized into regression and generative approaches.
Regression models \cite{potlapalli2306promptir,zhang2023ingredient,zhang2025perceive,zeng2025vision} often introduce unified frameworks to deal with diverse degradations, such as the prompt-based restoration strategy \cite{potlapalli2306promptir}, utilizing vision foundation models to extract specific degradation features \cite{zhang2025perceive}, adopting vision-language models for degradation classification \cite{zeng2025vision} and so on.
These models, though learning highly pixel-specific mappings from degraded inputs to clean labels, often fail to produce visually appealing results due to their tendency to create overly smooth, perceptually poor reconstructions.

Generative approaches, such as diffusion models \cite{miao2023dds2m,pang2024hir,zeng2024unmixing, pan2024diffsci, 10713101} and Autoregressive (AR) models \cite{zhan2022auto,han2025infinity}, are explicitly designed to produce high-fidelity outputs with convincing visual quality and rich texture.
By employing the denoising process of diffusion models (or the sequential prediction of AR models), these models learn the underlying distribution of real, clean images, enabling them to synthesize fine details and realistic components. 
However, these models face challenges in practical image restoration: (1) diffusion models suffer from extremely high computational cost and slow sampling speed; (2) traditional AR models involve sequential sampling for images, which constrains their modeling of spatial and spectral dependencies.
\textit{These methods often lead to the generation of unfaithful results and are hard for real-world fast applications.}

Currently, another type of AR model, Visual Autoregressive Modeling (VAR) \cite{tian2024visual}, has emerged as a significant paradigm shift.
VAR utilizes a multi-scale Vector Quantized Variational AutoEncoder (VQVAE) \cite{van2017neural} to establish a hierarchical representation, which redefines image generation as the next-scale prediction (see Figure \ref{fig:highlight}), achieving significantly reduced inference computation cost and high fidelity compared to current generative models.
Despite its success in High-Quality (HQ) image generation, the application of VAR to image restoration, particularly to all-in-one HSI restoration, remains largely an untapped potential and faces the following three critical challenges: \textit{(1) Semantic Alignment: how can the model effectively ensure semantic-level feature alignment between HQ labels and the Low-Quality (LQ) conditions? (2) Multi-Degradation Management: how can the VAR-based model discriminate among and manage the latents of multiple, diverse degradation patterns within a unified VAR framework? (3) Spatial-Spectral Fidelity: how can the process simultaneously guarantee spatial-spectral structure of HSIs when decoding?}

In this paper, we first propose HSI-VAR, rethinking HSI restoration as a visual autoregressive generation problem which progressively models the spectral and spatial dependencies in a scale-wise manner.
\textbf{Firstly}, to improve the feature generation process, we propose the semantic alignment strategy between the HQ latents and the conditions, reducing the domain discrepancy caused by multiple degradations;
\textbf{Secondly}, to effectively manage different degradations, we adopt the Degradation-Aware Guidance (DAG), guiding the generation of more degradation-specific features;
\textbf{Thirdly}, to guarantee the accurate and faithful restoration, we introduce the Spatial-Spectral Adaptation (SSA) and jointly train it with the VQVAE decoder, preserving the latents to ensure strict spatial and spectral adherence.
Contributions are summarized as follows:
\begin{itemize}
    \item We propose HSI-VAR, first rethinking the HSI restoration as an visual autoregressive generation problem which progressively models spectral and spatial dependencies rather than global reconstruction.
    \item To enhance the VAR framework for all-in-one HSI restoration, we introduce three core components: alignment strategy to mitigate domain discrepancy, DAG for robust degradation management and SSA to precisely preserve spatial and spectral structure.
    \item Extensive experiments demonstrate that our HSI-VAR yields perceptually preferable results and much lower inference computation cost among compared approaches.
\end{itemize}

%% file: sec/2_relatedworks.tex
\section{Related Works}
\label{sec:relatedworks}
\subsection{Regression Methods for Restoration}
\label{sec:aioir_nongen}
PromptIR \cite{potlapalli2306promptir} employs a prompt generation module that integrates task-specific prompts into the RGB restoration model.
VLU-Net \cite{zeng2025vision} adopts the vision-language models to automatically select the degradation matrices in a multi-modal vector space.
PromptHSI \cite{lee2024prompthsi} first introduces all-in-one strategies into HSI restoration with vision-language adaptation.
MP-HSIR \cite{wu2025mp} adopts multiple prompts to guide the de-degradation process with visual, textual and spectral embeddings.
\textit{While these methods enjoy high scores on pixel-level metrics, they learn overly smooth results due to the pixel-specific mapping, lacking the perceptual fidelity required for real-world HSIs.}

\begin{figure*}[t]
    \centering
    \includegraphics[width=1\linewidth]{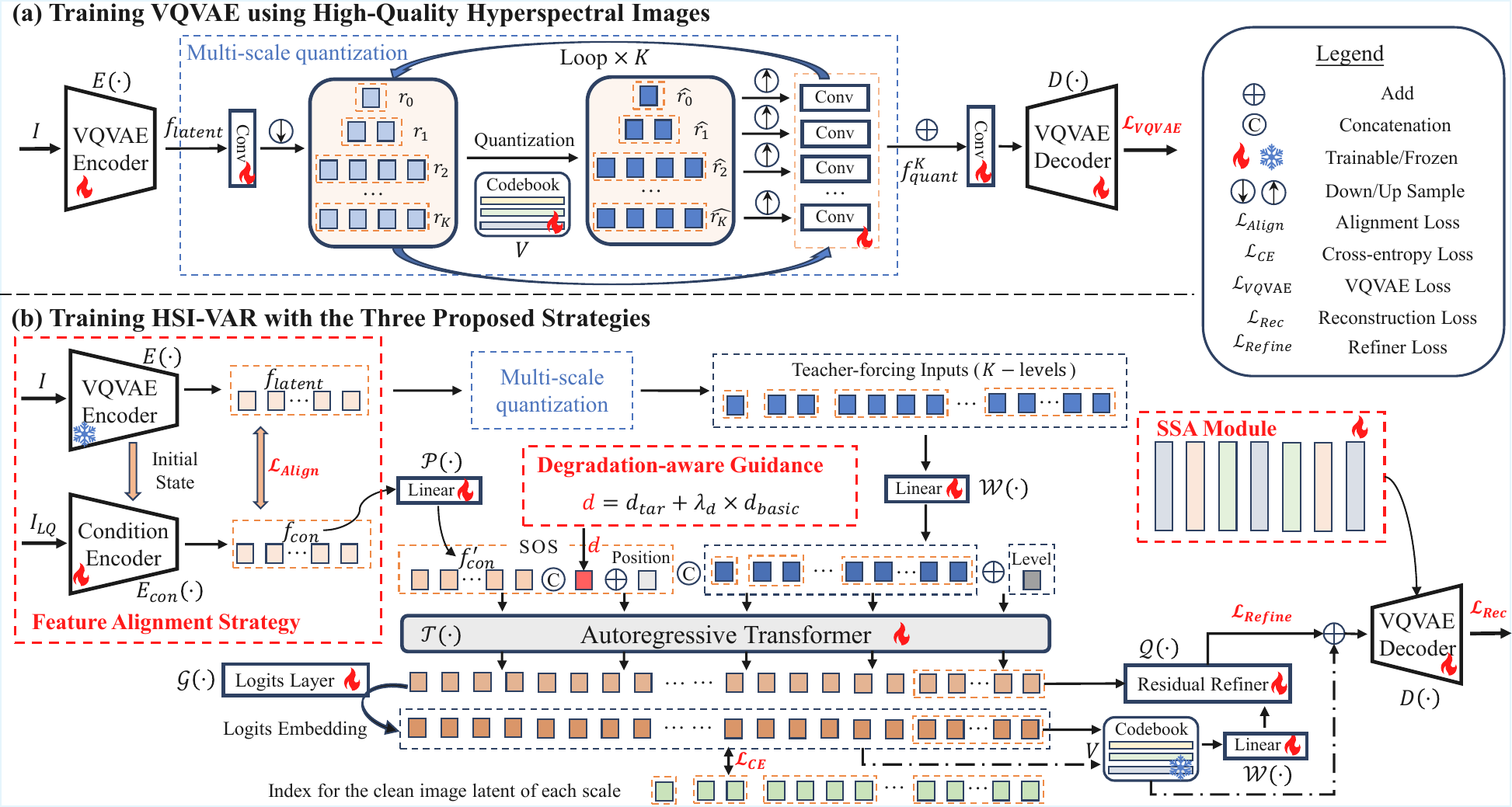}
    \vspace{-5mm}
    \caption{Illustration of the architecture of our HSI-VAR. (a) Structure of the multi-scale VQVAE. (b) HSI-VAR with the proposed three core components: feature alignment strategy, degradation-aware guidance and the spatial-spectral adaptation.}
    \vspace{-3mm}
    \label{fig:main}
\end{figure*}

\subsection{Diffusion-based Methods for Restoration}
\label{sec:aioir_gen_diff}
Diff-Plugin \cite{liu2024diff} trains task-specific plugins and injects these task priors into the latent diffusion model for RGB restoration.
PSRSCI \cite{zeng2025spectral} proposes a decomposition strategy to promote the HSI to RGB-like inputs and then finetunes the diffusion with the refined images.
AutoDIR \cite{jiang2024autodir} adopts a contrastive language-image pre-training model to generate degradation information guiding diffusion process.
DDS2M \cite{miao2023dds2m} uses a self-supervised strategy with diffusion models and HIR-Diff \cite{pang2024hir} develops the HSIs to three-channel inputs via singular value thresholding algorithm and then processes the diffusion steps.
\textit{Though diffusion-based methods have achieved good perceptual quality and sample fidelity, they are fundamentally constrained by high computational complexity and slow sampling speed.}

\subsection{AR/VAR-based Methods for Vision}
\label{sec:aioir_gen_ar}
AR models generate sequences by predicting the next element conditioned on preceding ones.
Following the advent of the vector quantization (VQVAE \cite{van2017neural}, VQGAN \cite{esser2021taming}), AR methods shifts towards handling discrete latent-level tokens (ImageGPT \cite{chen2020generative}, RQ-Transformer \cite{lee2022autoregressive}, LAR-SR \cite{guo2022lar}).
However, this token-flattening process disrupts the intrinsic spatial structure of images.
More recently, VAR \cite{tian2024visual} models the latent images into a coarse-to-fine multi-scale pattern, and then VAR-based methods such as VarFormer \cite{wang2025navigating}, HART \cite{tang2024hart}, VARSR \cite{qu2025visual} and RestoreVAR \cite{rajagopalan2025restorevar} have successfully explored its generative capabilities.
To the best of our knowledge, only two generative works have investigated the VAR in image restoration: VARSR \cite{qu2025visual} targets the super-resolution and RestoreVAR \cite{rajagopalan2025restorevar} first directly trains VAR for all-in-one RGB restoration.
\textit{Crucially, none of these methods incorporates the spectral priors inherent in HSIs, nor do they offer a dedicated solution for handling the multi-degradation problem within the VAR paradigm.}

%% file: sec/3_methods.tex
\section{Methods}
\subsection{Preliminaries: Visual Autoregressive Modeling}
\label{preliminaries_var}
VAR is a novel autoregressive class-conditioned image generation method which uses a decoder-only transformer architecture for next-scale prediction.

\subsubsection{Multi-scale VQVAE}
The VQVAE encoder $E$ initiates the process by transforming the input HQ image  $I \in \mathbb{R}^{H \times W \times C}$ into a continuous latent representation $f_{latent} \in \mathbb{R}^{h_K \times w_K \times c}$.
Instead of employing direct quantization on latent feature, the architecture implements a multi-scale residual quantization scheme utilizing a shared codebook $V \in \mathbb{R}^{M \times c}$ across $K$ spatial scales.
The procedure begins with the initialization of the residual feature $f_{res} \in \mathbb{R}^{h_K \times w_K \times c}$ and the accumulated quantized reconstruction $f_{quant} \in \mathbb{R}^{h_K \times w_K \times c}$, which are:
\begin{equation}
 f_{res}^{(0)} := f_{latent}, \
 f_{quant}^{(0)} := 0.
\end{equation}

This process iterates sequentially through each scale $k=1, \dots, K$.
At each step, the residual feature from the previous scale $f_{res}^{(k-1)}$ is first downsampled to the current resolution $h_{k} \times h_{k}$, and then quantized against the codebook $V$ (which means to find the nearest embedding from the codebook $V$), thereby determining the token map $r_{k} \in [V]^{h_{k} \times h_{k}}$.
This is formally represented as:
\begin{equation}
r_{k} := \operatorname{Quantize}_{V}\left(\operatorname{Downsample}\left(f_{res}^{(k-1)}, k\right)\right).
\end{equation}
Subsequently, the quantized token map $r_k \in \mathbb{R}^{h_{k} \times h_{k} \times c}$ is processed through upsampling and refinement by a convolutional module $\operatorname{Conv}_{k}$, generating a scale-specific residual reconstruction $h_k \in \mathbb{R}^{h_{K} \times w_{K} \times c}$ with full latent resolution.
These can be fomulated as:
\begin{equation}
h_{k} := \operatorname{Conv}_{k}\left(\operatorname{Upsample}\left(r_k, K\right)\right).
\end{equation}
This residual reconstruction $h_{k}$ serves to augment the accumulated quantized feature $f_{quant}^{(k)}$ and define the subsequent residual feature $f_{res}^{(k)}$ for the next iteration:
\begin{equation}
f_{quant}^{(k)} := f_{quant}^{(k-1)} + h_{k}, \
f_{res}^{(k)} := f_{res}^{(k-1)} - h_{k}.
\end{equation}
Upon completion of the $K$-th iterations, this hierarchical procedure yields a comprehensive set of token maps $\{r_{1}, r_{2}, \dots, r_{K}\}$, collectively encoding the image information across progressively increasing levels of detail.

\subsubsection{Next-scale Prediction}
VAR employs a teacher-forcing strategy to train its transformer blocks, utilizing the ground-truth token maps $\{r_{1}, r_{2}, \dots, r_{K}\}$ from the pre-trained multi-scale VQVAE to predict the next scale.
The transformer's input sequence is constructed by concatenating several components:
(1) A Start-Of-Sequence (SOS) token $\mathbf{s}_{{sos}}$, which is typically derived from the conditional embeddings $e$, is prepended to the sequence.
(2) The accumulated reconstruction features $f_{quant}^{(k)}$ at each scale are flattened into tokens, concatenated together and projected through the word embedding layer $\mathcal{W}$ to form the input sequence.
(3) VAR contains the some basic embeddings such as the position embedding $\mathbf{p}$ and the scale level embedding $\mathbf{l}$.
For training, the transformer's input sequence can be formulated as:
\begin{equation}
\label{input_sequence}
    s = \left[ \mathbf{s}_{{sos}} \parallel \mathcal{W}(f_{{quant}}^{(1)} \parallel \cdots \parallel f_{{quant}}^{(K)}) + \mathbf{p} + \mathbf{l} \right],
\end{equation}
where $\parallel$ denotes the token concatenation.
A block-wise causal attention mask is applied to ensure the predictions for current scale $r_k$ only attend to the information from all previous scales and the condition $e$.
VAR is trained to minimize the cross-entropy loss between the logits and the ground-truth index maps, effectively modeling the joint likelihood:
\begin{equation}
p(r_{1}, r_{2}, \dots, r_{K}) = \prod_{k=1}^{K} p(r_{k} | e, r_{1}, r_{2}, \dots, r_{k-1}).
\end{equation}

\subsection{The Proposed HSI-VAR Framework}
To better harness VAR's highly efficient generation capabilities for restoration, our HSI-VAR introduces basic modifications of VAR in Section \ref{ar_blocks}.
To address three critical challenges mentioned in Section \ref{sec:intro}, we propose the specific solutions in Section \ref{feature_align}, \ref{dacfg} and \ref{s3a_module} (details are presented in Figure \ref{fig:main} and Figure \ref{fig:main_test}).

\subsubsection{VAR Architecture for Restoration}
\label{ar_blocks}
Based on the architecture detailed in Section \ref{preliminaries_var} for class-conditioned generation, we introduce several basic modifications to adapt VAR for the restoration task.
(1) Conditional Prefix Tokens: instead of using class label embeddings $e$, we utilize the degraded LQ inputs $I_{LQ}$ to generate the conditional embeddings by the conditional encoder $E_{con}$ as prefix tokens. 
This allows the tokens being predicted to condition not only on preceding scale tokens but also on the robust LQ information, fundamentally guiding the restoration process;
(2) Positional Encoding: we replace the standard absolute position embeddings with 2D Rotary Positional Embeddings (RoPE) \cite{su2024roformer}.
RoPE inherently supports resolution scaling and significantly enhances the structural fidelity required for high-fidelity hyperspectral image restoration;
(3) Removing AdaLN Layers: the AdaLN layers are removed to reduce the extra parameters, streamlining the model architecture;
(4) High-Fidelity Refiner: we incorporate a refiner $\mathcal{Q}$ based on the NAFNet \cite{chen2022simple}, which compensates for the quantization loss $f_{res}^{K}$ in the latent space. The loss is:
\begin{equation}
    \mathcal{L}_{Refiner} = \|\mathcal{Q}(\mathcal{W}(r_K),z)-f_{res}^{K}\|_1,
\end{equation}
where $z$ is direct last scale output from the block-wise autoregressive transformer $\mathcal{T}$.

\subsubsection{Alignment between HQ Images and Conditions}
\label{feature_align}
\begin{figure}
    \centering
    \hspace{-1mm}
    \includegraphics[width=0.95\linewidth]{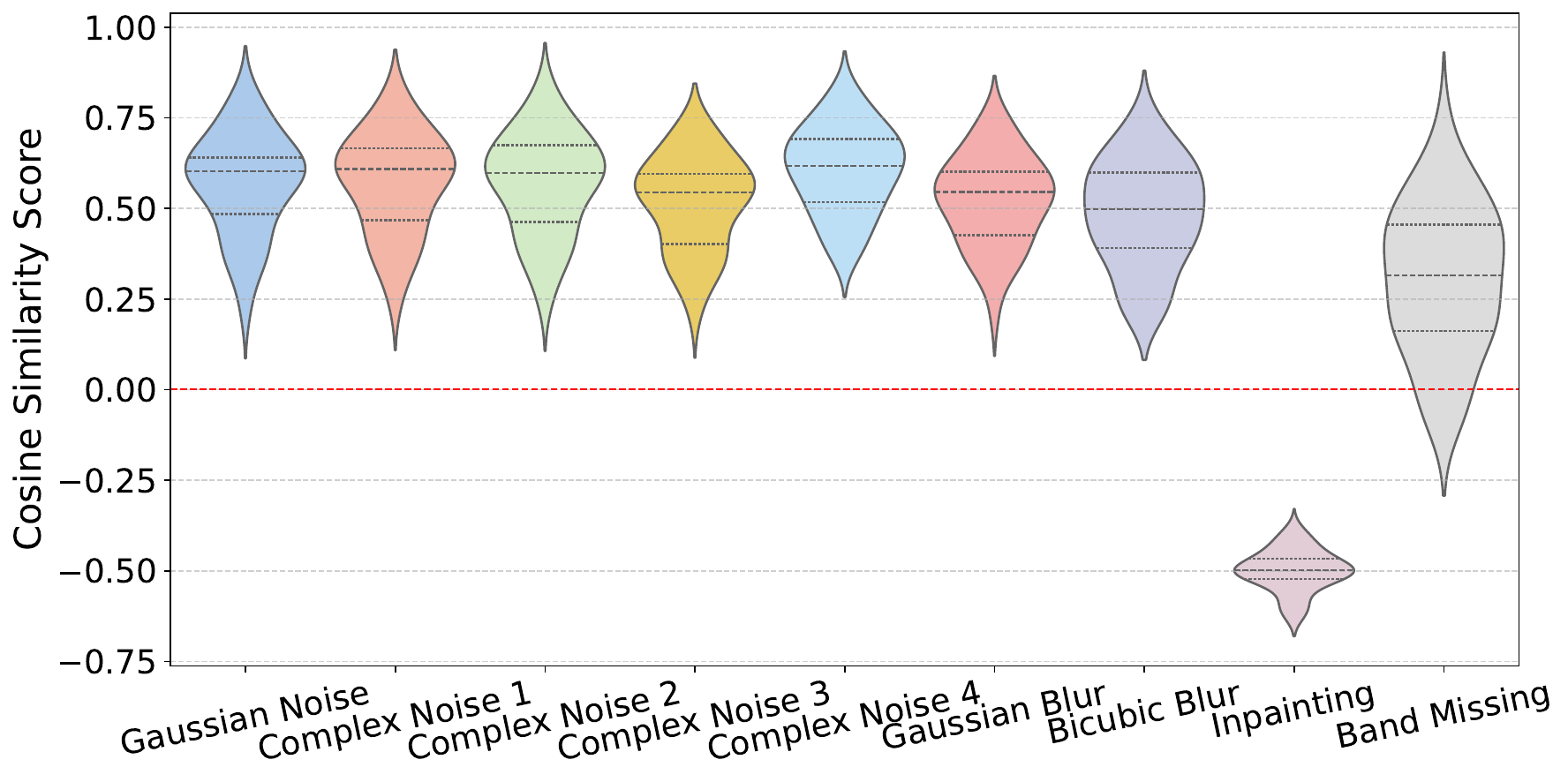}
    \vspace{-2mm}
    \caption{Motivation for Alignment: Pre-trained encoder pocesses similar latents for degraded inputs even without extra training.}
    \vspace{-4mm}
    \label{fig:similarity}
\end{figure}

\textbf{Motivation.} Current VAR-based methods often apply ControlNet \cite{zhang2023adding} as the condition encoder $E_{con}$ to add controls for generative tasks.
While due to the impact of multiple and diverse degradations, there exists a large feature gap in the discrete latent space between the HQ encoded latents and the conditional embeddings from $E_{con}$, leading to the loss of content-aware textures for the restoration.
Inspired by the pre-trained VQVAE encoder $E$ which naturally poses the HQ information, we analyze the similarity scores of the latents encoded from both HQ inputs and the corresponding degraded HSIs in Figure \ref{fig:similarity}, exploring the possibility of an alternative to align the features.
From Figure \ref{fig:similarity}, the similarity scores for nearly all degraded latents reach high, denoting that the latent semantic encoding property of $E$ also works for degraded inputs even without extra training.
Specifically, the state of complete irrelevance of the inpainting degradation is often due to the loss of the large-area pixel-level information.
Based on above analysis, we propose the following alignment strategy, which aims to minimize feature inconsistency between latents from pre-trained VQVAE encoder $E$ and the conditional encoder $E_{con}$.

\begin{figure*}
    \centering
    \includegraphics[width=1\linewidth]{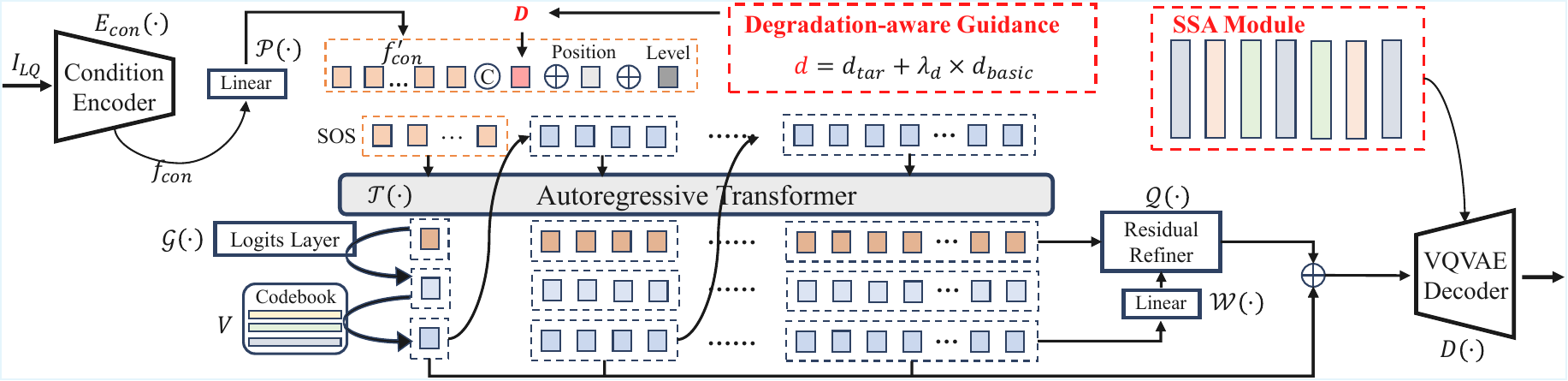}
    \vspace{-5mm}
    \caption{Illustration of the inference pipeline of our HSI-VAR, which incorporates modules and processes features scale by scale.}
    \vspace{-3mm}
    \label{fig:main_test}
\end{figure*}

\noindent \textbf{Details.} This strategy (see the left of Figure \ref{fig:main} (b)):
(1) first adopts and initializes the conditional encoder $E_{con}$ using the pre-trained HQ encoder $E$:
\begin{equation}
E_{con} := E,
\end{equation}
(2) then refines $E_{con}$ by minimizing an alignment loss $\mathcal{L}_{Align}$, ensuring that the feature space of the LQ condition aligns with that of the HQ features.
The loss is defined as the expected squared L2 norm between the features:
\begin{equation}
\mathcal{L}_{Align} =  \left\| E_{con}(I_{LQ}) - E(I_{HQ}) \right\|_{2}^{2},
\end{equation}
where $I_{HQ}$ and $I_{LQ}$ denote the HQ input and the corresponding degraded LQ HSI.
This strategy boosts the conditional representation and effectively guides the generation for superior HSIs.
The encoded features $f_{con} = E_{con}(I_{LQ})$ are then reshaped and projected through a linear layer $\mathcal{P}$ as $f_{con}^{'}$, which is finally the image-based condition in $\textbf{s}_{sos}$.

\subsubsection{DAG for Multi-degradation Management}
\label{dacfg}
\begin{figure}
        \hspace{-3mm}
        \begin{tabular}{cc}
        \centering
            \hspace{-1mm}\includegraphics[width=0.5\linewidth]{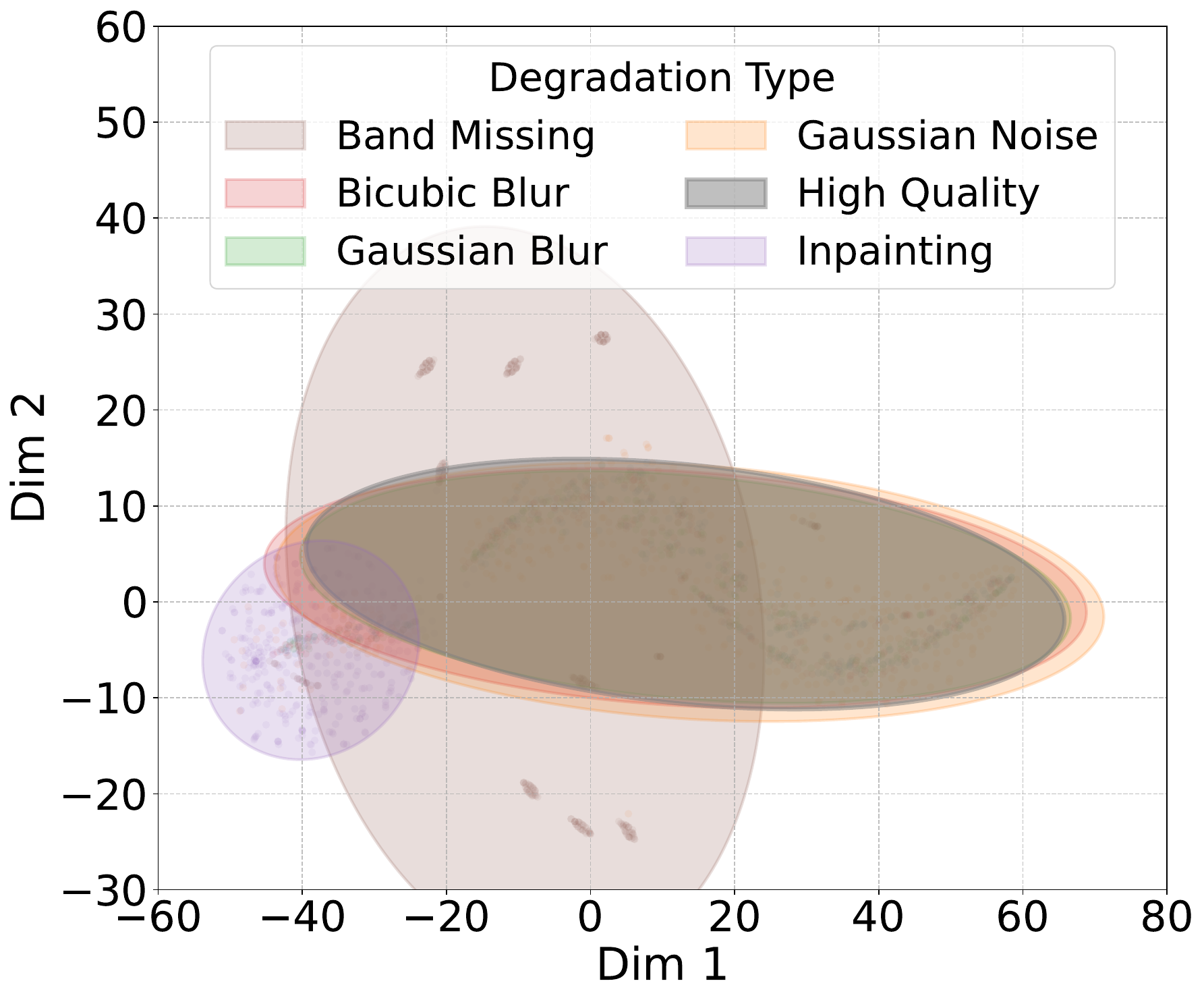} &\hspace{-4mm}\raisebox{-1pt}{\includegraphics[width=0.5\linewidth]{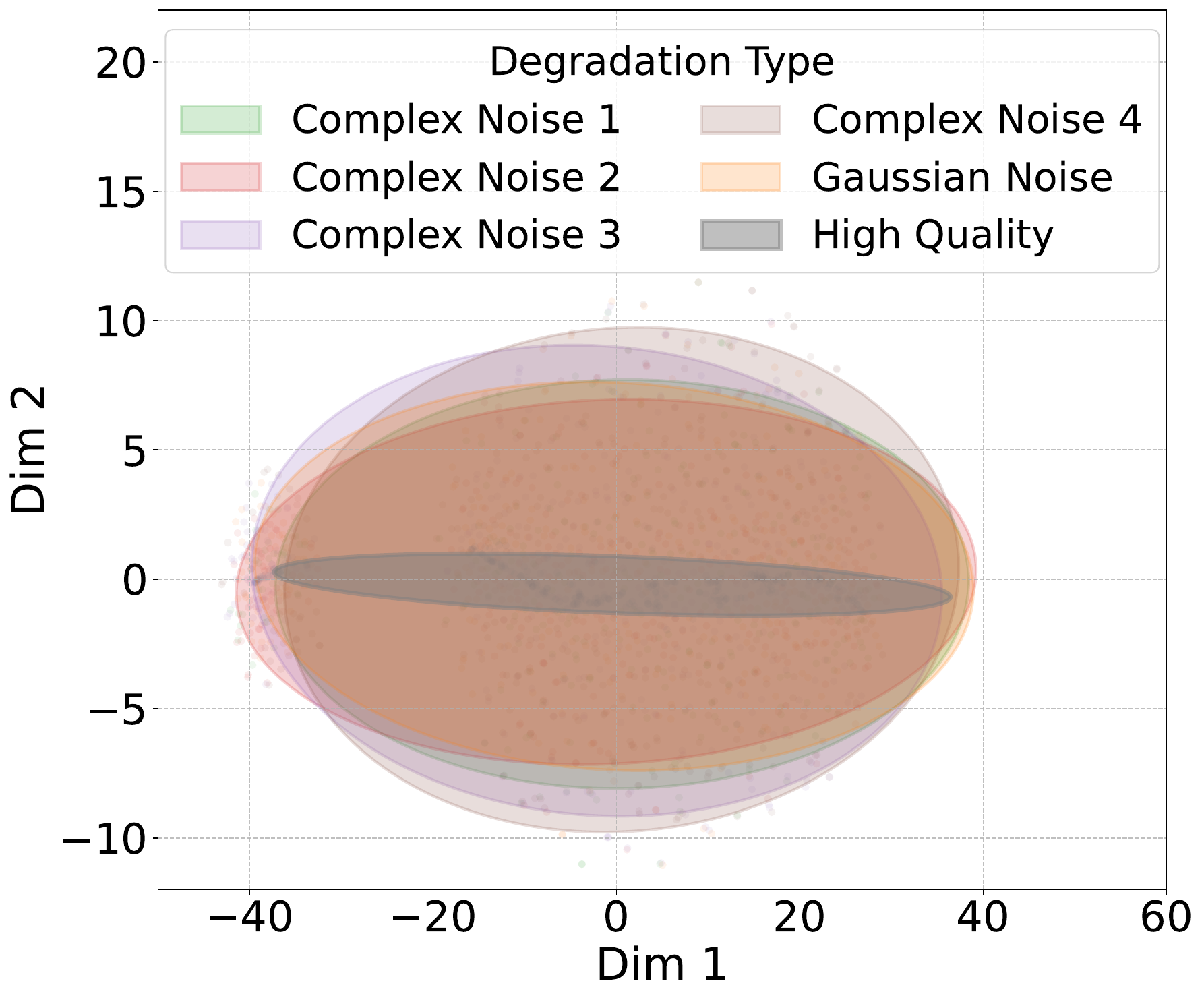}}\\
        \end{tabular}
        \vspace{-4mm}
    \caption{Motivation for DAG: Different degradations have not only the specificities but also the commonalities.}
    \vspace{-5mm}
    \label{fig:motivation_dacfg}
\end{figure}

\textbf{Motivation.} A distribution analysis among $N$ degradations in the HSI restoration in Figure \ref{fig:motivation_dacfg} reveals that these degradations exhibit not only unique characteristics but also shared properties, thus a mixed modeling strategy should be considered.
And current Classifier-Free Guidance (CFG) \cite{ho2022classifier} incurs additional computational overhead during inference as it requires computing predictions for both the conditional $e_{con}$ and unconditional $e_{un}$ embeddings, which is:
\begin{equation}
\label{eq:output_guidance}
\tilde{logits_{e}} = \mathcal{G}(\mathcal{T}(e_{un})) + \lambda_{e} \times \left(\mathcal{G}(\mathcal{T}(e_{con})) - \mathcal{G}(\mathcal{T}(e_{un}))\right),
\end{equation}
where $\tilde{logits_{e}}$ represents the logits after CFG, $\lambda_{e}$ denotes the guidance scale and $\mathcal{G}$ is the logits layer.
Noted that finally this strategy should tune $\lambda_{e}$ for better guided results.

\noindent \textbf{Details.} 
Therefore we propose DAG, which adopts $N$ degradation-specific embeddings ($d_{1}, d_2, \cdots, d_{N}$), $N$ corresponding guidance scale ($\lambda_1, \lambda_2, \cdots, \lambda_N$) and one basical degradation embedding $d_{basic}$ in the embedding space.
Specifically for training, despite the encoded and aligned (in Section \ref{feature_align}) LQ conditions $f_{con}^{'}$ as the general image-based condition, the generation process is also conditioned by the linear combination with one selected specific target degradation embedding $d_{tar} := d_{i}$, the corresponding guidance scale $\lambda_d:=\lambda_i$ and the basical degradation embedding $d_{basic}$, jointly for task-specific and general awareness.
For each iteration, this mixed degradation embedding is concatenated with $f_{con}^{'}$ for causal scale-wise modeling.
Related derivations are shown below:
\begin{equation}
\label{dag}
    d = d_{tar} + \lambda_d \times d_{basic},
\end{equation}
\begin{equation}
    \tilde{logits_{d}} = \mathcal{G}(\mathcal{T}(f_{con}^{'}, d)).
\end{equation}
Noted that the guidance scale $\lambda_{d}$ is also learnable as a parameter, automatically governing the balance between specific degradation and generalized image fidelity in training.
In the test phase, since we combine the conditions in the embedding space, the generation only needs the inference in one time.
Thus our DAG proposes a dedicated solution for multi-degradation management within the VAR paradigm, which not only halves the computational cost but also eliminates the need to tune the CFG ratio as a hyper-parameter.

\begin{table*}[htbp]
    \centering
    \caption{All-in-one HSI restoration performance comparison with regression methods.}
    \vspace{-2mm}
    \label{tab:comparison_non_generative}
    \setlength{\tabcolsep}{2.5pt}
    \resizebox{\textwidth}{!}{%
    \begin{tabular}{ccc c c c c c c c c c c c c c c}
        \toprule
        \multirow{3}{*}{\textbf{Dataset}} & \multirow{3}{*}{\textbf{Method}} & \multirow{3}{*}{\textbf{Ref.}} & \multicolumn{2}{c}{\textbf{i.i.d G-Denoise}} & \multicolumn{2}{c}{\textbf{C-Denoise}} & \multicolumn{2}{c}{\textbf{G-Deblur}} & \multicolumn{2}{c}{\textbf{SR}} & \multicolumn{2}{c}{\textbf{IP}} & \multicolumn{2}{c}{\textbf{BC}} & \multicolumn{2}{c}{\textbf{Average}} \\
        \cline{4-17} 
        & & & \multicolumn{2}{c}{\small ($\sigma=\{30, 50, 70\}$)} & \multicolumn{2}{c}{\small (Four Cases)} & \multicolumn{2}{c}{\small (Radius = $9, 15, 21$)} & \multicolumn{2}{c}{\small (Scale = $2, 4, 8$)} & \multicolumn{2}{c}{\small (Mask Rate = $0.7, 0.8, 0.9$)} & \multicolumn{2}{c}{\small (Rate = $0.1, 0.2, 0.3$)} & \multicolumn{2}{c}{\small (All Degradations)} \\
        \cline{4-17} 
        & & & \small \textbf{CLIP-IQA}$\uparrow$ & \small \textbf{MANIQA}$\uparrow$ & \small \textbf{CLIP-IQA}$\uparrow$ & \small \textbf{MANIQA}$\uparrow$ & \small \textbf{CLIP-IQA}$\uparrow$ & \small \textbf{MANIQA}$\uparrow$ & \small \textbf{CLIP-IQA}$\uparrow$ & \small \textbf{MANIQA}$\uparrow$ & \small \textbf{CLIP-IQA}$\uparrow$ & \small \textbf{MANIQA}$\uparrow$ & \small \textbf{CLIP-IQA}$\uparrow$ & \small \textbf{MANIQA}$\uparrow$ & \small \textbf{CLIP-IQA}$\uparrow$ & \small \textbf{MANIQA}$\uparrow$ \\
        \midrule
        \multirow{3}{*}{\textbf{\textit{ICVL}}}
        & PromptIR & NeurIPS'23 & \large 0.266 & \large 0.203 & \large 0.265 & \large 0.202 & \large 0.316 & \large 0.206 & \large 0.307 & \large 0.212 & \large 0.283 & \large 0.202 & \large 0.274 & \large 0.202 & \large 0.285 & \large 0.205 \\
        & VLUNet & CVPR'25 & \large 0.309 & \large 0.208 & \large 0.293 & \large 0.208 & \large 0.311 & \large 0.206 & \large 0.315 & \large 0.213 & \large 0.285 & \large 0.202 & \large 0.271 & \large 0.202 & \large 0.297 & \large 0.207 \\
        \rowcolor{lightgray!20}
        \cellcolor{white}
        & HSI-VAR & Ours & \large \textbf{0.347} & \large \textbf{0.216} & \large \textbf{0.347} & \large \textbf{0.215} & \large \textbf{0.356} & \large \textbf{0.208} & \large \textbf{0.349} & \large \textbf{0.213} & \large \textbf{0.355} & \large \textbf{0.218} & \large \textbf{0.359} & \large \textbf{0.216} & \large \textbf{0.352} & \large \textbf{0.210} \\
        \midrule
        \multirow{3}{*}{\textbf{\textit{ARAD}}}
        & PromptIR & NeurIPS'23 & \large 0.313 & \large 0.216 & \large 0.315 & \large 0.216 & \large \textbf{0.332} & \large \textbf{0.237} & \large 0.306 & \large 0.209 & \large \textbf{0.331} & \large 0.225 & \large 0.320 & \large 0.220 & \large 0.320 & \large 0.221 \\
        & VLUNet & CVPR'25 & \large 0.325 & \large 0.214 & \large 0.328 & \large 0.228 & \large 0.325 & \large 0.226 & \large 0.302 & \large 0.209 & \large 0.323 & \large 0.224 & \large 0.320 & \large 0.220 & \large 0.319 & \large 0.218 \\
        \rowcolor{lightgray!20}
        \cellcolor{white}
        & HSI-VAR & Ours & \large \textbf{0.331} & \large \textbf{0.230} & \large \textbf{0.329} & \large \textbf{0.230} & \large 0.322 & \large 0.197 & \large \textbf{0.317} & \large \textbf{0.213} & \large 0.328 & \large \textbf{0.232} & \large \textbf{0.328} & \large \textbf{0.234} & \large \textbf{0.326} & \large \textbf{0.225} \\
        \bottomrule
    \end{tabular}
    }
    \vspace{-2mm}
\end{table*}

\begin{table*}[htbp]
    \centering
    \caption{All-in-one HSI restoration performance comparison with generative methods.}
    \vspace{-2mm}
    \setlength{\tabcolsep}{3pt}
    \label{tab:col_comparison}
    \resizebox{\textwidth}{!}{%
    \begin{tabular}{cccccccccc|ccccccc} 
        \toprule
        \multirow{3}{*}{\textbf{Degradation}} & \multirow{3}{*}{\textbf{Method}} & \multirow{3}{*}{\textbf{Ref.}} & \multicolumn{7}{c}{\textit{\textbf{ICVL}}} & \multicolumn{7}{c}{\textit{\textbf{ARAD}}} \\
        \cline{4-17}
        & & & \textbf{PSNR} $\uparrow$ & \textbf{SSIM} $\uparrow$ & \textbf{LPIPS} $\downarrow$ & \textbf{DISTS} $\downarrow$ & \textbf{CLIP-IQA} $\uparrow$ & \textbf{MUSIQ} $\uparrow$ & \textbf{MANIQA} $\uparrow$
        & \textbf{PSNR} $\uparrow$ & \textbf{SSIM} $\uparrow$ & \textbf{LPIPS} $\downarrow$ & \textbf{DISTS} $\downarrow$ & \textbf{CLIP-IQA} $\uparrow$ & \textbf{MUSIQ} $\uparrow$ & \textbf{MANIQA} $\uparrow$ \\
        \midrule
        
        \multirow{2}{*}{\shortstack{\textbf{i.i.d G-Denoise} \\ ($\sigma=\{ 30, 50, 70\}$)}}
        & PSRSCI & ICLR'25 & \large22.08 & \large.684 & \large.275 & \large.214 & \large.309 & \large32.96 & \large.239 & \large20.31& \large.613& \large.273& \large\underline{.220}& \large\textbf{.365}& \large\textbf{39.56}& \large\textbf{.256} \\
        & VARSR & ICML’25 & \large\textbf{34.45} & \large\underline{.893} & \large\underline{.227} & \large\underline{.205} & \large\underline{.343} & \large28.33 & \large\textbf{.207} & \large\underline{29.35} & \large\underline{.862} & \large\underline{.239} & \large.226 & \large.322 & \large31.31 & \large.220 \\
        \rowcolor{lightgray!20}
        \cellcolor{white}
        & HSI-VAR & Ours & \large\underline{33.94}& \large\textbf{.921}& \large\textbf{.193}& \large\textbf{.187}& \large\textbf{.347}& \large \underline{28.97}& \large\underline{.216} & \large\textbf{29.57}& \large\textbf{.870}& \large\textbf{.210}& \large\textbf{.206}& \large\underline{.331}& \large\underline{34.73}& \large\underline{.230} \\
        \bottomrule
        
        \multirow{2}{*}{\shortstack{\textbf{C-Denoise} \\ (Four Cases)}}
        & PSRSCI & ICLR'25 & \large21.87 & \large.686 & \large.283 & \large.217 & \large.310 & \large\textbf{32.42} & \large\textbf{.240} & \large20.33& \large.612& \large.285& \large\underline{.225}& \large\textbf{.362}& \large\textbf{39.47}& \large.\textbf{256}\\
        & VARSR & ICML’25 & \large\textbf{33.56} & \large \underline{.895} & \large\underline{.224} & \large\underline{.212} & \large\textbf{.349} &	\large27.77 & \large.205 & \large\underline{28.59} & \large\underline{.844} & \large\underline{.249} & \large.234 & \large.324 & \large30.91 & \large.219\\
        \rowcolor{lightgray!20}
        \cellcolor{white}
        & HSI-VAR & Ours & \large\underline{33.46}& \large\textbf{.918}& \large\textbf{.197}& \large\textbf{.189}& \large\underline{.348}& \large\underline{28.81}& \large\underline{.215} & \large\textbf{29.41}& \large\textbf{.867}& \large\textbf{.211}& \large\textbf{.207}& \large\underline{.329}& \large\underline{34.74}& \large\underline{.230}\\
        \bottomrule
        
        \multirow{2}{*}{\shortstack{\textbf{G-Deblur} \\ (Radius = $9, 15, 21$)}}
        & PSRSCI & ICLR'25 & \large\underline{29.51} & \large\underline{.885} & \large\textbf{.130} & \large\textbf{.142} & \large.337 & \large\underline{36.69} & \large\underline{.230} & \large\underline{26.37}& \large\underline{.824}& \large\textbf{.159}& \large\textbf{.158}& \large\textbf{.379}& \large\underline{37.28}& \large\textbf{.243}\\
        & VARSR & ICML’25 & \large25.47 & \large.639 & \large.411 & \large.298 & \large\textbf{.379} & \large\textbf{37.87} & \large.250 & \large20.42 & \large.539 & \large.445 & \large.314 & \large\underline{.357} & \large\textbf{40.50} & \large\underline{.206}\\
        \rowcolor{lightgray!20}
        \cellcolor{white}
        & HSI-VAR & Ours & \large\textbf{33.18}& \large\textbf{.902}& \large\underline{.282}& \large\underline{.229}& \large\underline{.356}& \large22.85& \large.209 & \large\textbf{29.78}& \large\textbf{.857}& \large\underline{.305}& \large\underline{.255}& \large.322& \large26.40& \large.197\\
        \bottomrule
        
        \multirow{2}{*}{\shortstack{\textbf{SR} \\ (Scale = $2, 4, 8$))}}
        & PSRSCI & ICLR'25 & \large28.18 & \large\underline{.841} & \large\textbf{.158} & \large\textbf{.164} & \large.310 & \large\underline{28.50} & \large\textbf{.219} & \large25.44& \large.767& \large\textbf{.200}& \large\textbf{.188}& \large\textbf{.345}& \large\textbf{34.65}& \large\textbf{.231}\\
        & VARSR & ICML’25 & \large\underline{29.18} & \large.789 & \large.299 & \large.245 & \large\textbf{.371} &\large\textbf{29.38} & \large\underline{.214} & \large\textbf{28.77}& \large\textbf{.838}& \large.252& \large.232& \large.317& \large31.34& \large.213\\
        \rowcolor{lightgray!20}
        \cellcolor{white}
        & HSI-VAR & Ours & \large\textbf{32.65}& \large\textbf{.891}& \large\underline{.231}& \large\underline{.209}& \large\underline{.349}& \large26.56& \large.213& \large\underline{28.53} & \large\underline{.835} & \large\underline{.251} & \large\underline{.229} & \large.311 & \large\underline{31.71} & \large\underline{.215} \\
        \bottomrule

        \multirow{2}{*}{\shortstack{\textbf{IP} \\ (Mask Rate = $0.7, 0.8, 0.9$)}}
        & PSRSCI & ICLR'25 & \large22.30 & \large.719 & \large\underline{.253} & \large\underline{.222} & \large.345 & \large\underline{29.42} & \large\textbf{.231} & \large23.08& \large.716& \large\underline{.263}& \large\underline{.233}& \large\textbf{.365}& \large\underline{34.69}& \large\underline{.231}\\
        & VARSR & ICML’25 & \large\underline{24.20} & \large\underline{.787} & \large.314 & \large.275 & \large\textbf{.386} &\large25.69 & \large.206 & \large\underline{24.09} & \large\underline{.760} & \large.304 & \large.280 & \large\underline{.347} & \large28.12 & \large.215 \\
        \rowcolor{lightgray!20}
        \cellcolor{white}
        & HSI-VAR & Ours & \large\textbf{34.16}& \large\textbf{.930}& \large\textbf{.176}& \large\textbf{.176}& \large\underline{.355}& \large\underline{29.36}& \large\underline{.218} & \large\textbf{29.05}& \large\textbf{.872}& \large\textbf{.202}& \large\textbf{.197}& \large.328& \large\textbf{35.46}& \large\textbf{.232}\\
        \bottomrule

        \multirow{2}{*}{\shortstack{\textbf{BC} \\ (Rate = $0.1, 0.2, 0.3$)}}
        & PSRSCI & ICLR'25 & \large20.65 & \large.817 & \large\textbf{.119} & \large\textbf{.142} & \large.322 & \large\textbf{35.47} & \large\textbf{.222} & \large20.73& \large.759& \large\textbf{.146}& \large\textbf{.155}& \large\textbf{.406}& \large\textbf{45.51}& \large\textbf{.281}\\
        & VARSR & ICML’25 & \large\underline{29.89} &\large\underline{.922} & \large.177 & \large.183 & \large\underline{.352} & \large28.18 & \large.204 & \large\underline{26.67} & \large\textbf{.862} & \large.224 & \large.214 & \large.326 & \large32.29 & \large.221 \\
        \rowcolor{lightgray!20}
        \cellcolor{white}
        & HSI-VAR & Ours & \large\textbf{31.97}& \large\textbf{.924}& \large\underline{.168}& \large\underline{.174}& \large\textbf{.359}& \large\underline{29.82}& \large\underline{.216} & \large\textbf{27.70}& \large\underline{.861}& \large\underline{.204}& \large\underline{.198}& \large\underline{.328}& \large\underline{36.14}& \large\underline{.234}\\
        \bottomrule

        \multirow{2}{*}{\shortstack{\textbf{Average} \\ (All Degradations)}}
        & PSRSCI & ICLR'25 & \large 23.98 & \large.767 & \large\underline{.217} & \large\textbf{.185} & \large.322 & \large \textbf{32.57} & \large \textbf{.230} & \large 22.59 & \large .710 & \large \textbf{.225} & \large \textbf{.200} & \large \textbf{.370} & \large \textbf{38.58} & \large \textbf{.250} \\
        & VARSR & ICML’25 & \large\underline{29.46} & \large\underline{.838} & \large.275 & \large.236 & \large\textbf{.363} & \large\underline{29.54} & \large\underline{.215} & \large\underline{25.66} & \large\underline{.762} & \large.300 & \large.256 & \large\underline{.330} & \large32.46 & \large.214 \\
        \rowcolor{lightgray!20}
        \cellcolor{white}
        & HSI-VAR & Ours & \large\textbf{33.23} & \large\textbf{.915} & \large\textbf{.207} & \large\underline{.193} & \large\underline{.352} & \large27.79 & \large.210 & \large\textbf{29.06} & \large\textbf{.862} & \large\underline{.228} & \large\underline{.215} & \large.326 & \large\underline{33.22} & \large\underline{.225}\\
        \bottomrule
    \end{tabular}
    }
    \vspace{-4mm}
\end{table*}

\subsubsection{Structure Preservation with SSA Module}
\label{s3a_module}
\textbf{Motivation.} VQVAE is primarily trained to construct HQ semantic features (or codes) through multi-scale codebooks.
The encoder processes the pixel-level inputs to the condensed semantic codes, while the decoder needs to recover the precise original HSI from semantics with severely corrupted pixel information.
Thus this approach inevitably leads to the loss of the critical pixel-level features in the decoder and brings severely distorted reconstruction results, which goes against the natural requirements of HSI restoration: the model must preserve the spatial pixel-level information and robust consistency in the spectral dimension.

\noindent \textbf{Details.} To tackle this challenge, we introduce the SSA finetuning strategy (in Figure \ref{fig:main} (b)) to promote the decoder of the VQVAE with more pixel-level features.
We jointly train the decoder of the VQVAE and the SSA, bringing higher flexibility and better pixel information adaptation in the decoding phase.
Concretely, SSA consists of multiple spatial and spectral attention modules (Spa-A and Spe-A), which integrates both structures via weighted fusion for each decoder feature layer, which can be formulated as:
\begin{equation}
    f_i = \text{Spa-A}(f_i) + \sigma_i \times \text{Spe-A}(f_i),
\end{equation}
where for layer $i$, $f_i$ denotes the feature map and $\sigma_i$ is the weight scale initialized to zero.
The reconstruction loss is:
\begin{equation}
\label{recloss}
    \mathcal{L}_{Rec} = \|\tilde{I}-I \|_1 + \gamma \operatorname{SSIM}(\tilde{I}, I),
\end{equation}
where $\tilde{I}$ represents the final restored HSI and $\gamma$ denotes the hyper-parameter set to $0.2$.
This procedure effectively preserves spatial and spectral structures while learning finer details essential for superior reconstruction quality.

\subsubsection{Overall Training and Testing Pipeline}
\label{pipeline}
Training (in Figure \ref{fig:main}) contains the following stages.
First we finetune the pre-trained VQVAE with HQ HSIs, which are transported into the VQVAE encoder $E$, quantizer $\operatorname{Quantize}_{V}$ and VQVAE decoder $D$, obtaining the reconstructed HSIs.
Then we train the HSI-VAR.
$I_{LQ}$ is encoded by the $E_{con}$ and projected by $\mathcal{P}$ to $f_{con}^{'}$ as prefix tokens $\mathbf{s}_{{sos}}$.
The above tokens are concatenated or added with degradation embedding $d$ (as Eq.~\eqref{dag}), position embedding $\mathbf{p}$ and the scale-level embedding $\mathbf{l}$, forming the input sequence in Eq.~\eqref{input_sequence}.
After the scale-wise mask attention modeling with cross-entroy loss $\mathcal{L}_{CE}$, HSI-VAR adopts the hidden states and quantized latents to approximate the quantization loss $f_{res}^{K}$ via refiner $\mathcal{Q}$.
The training loss of HSI-VAR is:
\begin{equation}
    \mathcal{L}_{total} = \mathcal{L}_{CE} + \beta_1\mathcal{L}_{Refiner} + \beta_2\mathcal{L}_{Align},
\end{equation}
where $\mathcal{L}_{CE}$ denotes the cross-entroy loss, $\beta_1, \beta_2$ are hyper-parameters and separately set as $2.0,0.5$.
Finally we finetune the VQVAE decoder $D$ and the SSA module with the reconstruction loss in Eq.~\eqref{recloss}.

Testing process (in Figure \ref{fig:main_test}) is initialized with the SOS token $\mathbf{s}_{{sos}}$ derived from the $f_{con}^{'}$, $d$, $\mathbf{p}$ and $\mathbf{l}$.
Our HSI-VAR then proceeds to autoregressively predict each token map $r_k$ in a sequential, scale-by-scale manner.
Finally, after adding up the quantization loss via the refiner $\mathcal{Q}$, the latents are decoded by the VQVAE decoder $D$ and the SSA module.

%% file: sec/4_experiments.tex
\section{Experiments}

\subsection{Experimental Setup}
\label{experiments}

\noindent \textbf{Datasets.}
We utilize two datasets, \textbf{\textit{ICVL}} \cite{arad2016sparse} and \textbf{\textit{ARAD}} \cite{arad2022ntire}.
\textbf{\textit{ICVL}} contain 100 HSIs for training with $1392 \times 1300$ resolution and 50 images for testing with $512 \times 512$ resolution.
\textbf{\textit{ARAD}} includes 900 HSIs for training and 50 images for testing with $482 \times 512$ resolution.
Images are cropped into $256 \times 256 \times 31$ patches. 
Normalization is uniformly conducted across all datasets.
Degradations are the independent and identically distributed (i.i.d) Gaussian Denoising (i.i.d G-Denoise), Complex Denoising (C-Denoise), Gaussian Deblurring (G-Deblur), Super-Resolution (SR), Inpainting (IP) and Band Completion (BC). 
We present detailed settings in the supplementary materials.

\begin{figure*}
    \centering
    \includegraphics[width=\linewidth]{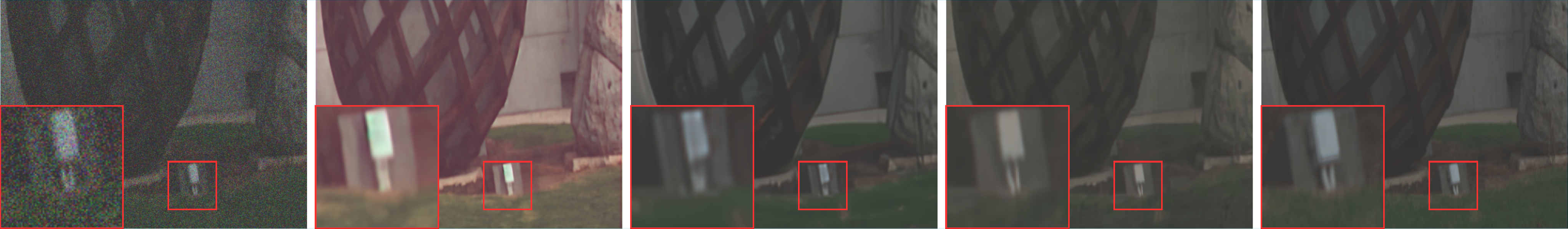}\\
    \vspace{0.1em}
    \includegraphics[width=\linewidth]{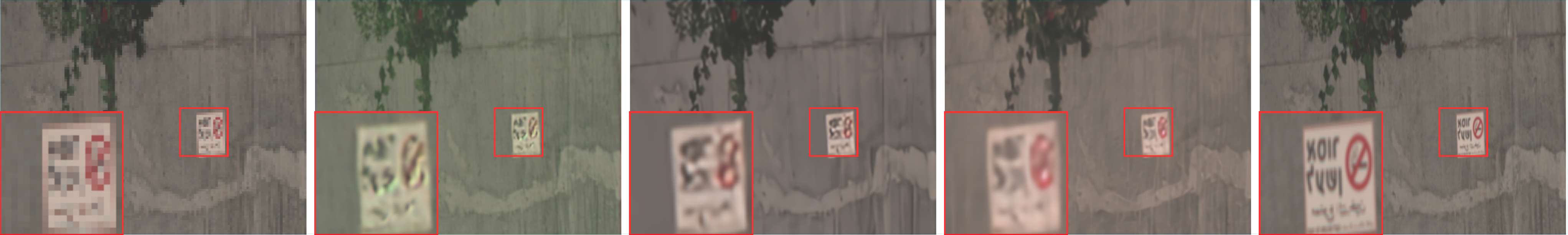}\\
    \vspace{0.1em}
    \includegraphics[width=\linewidth]{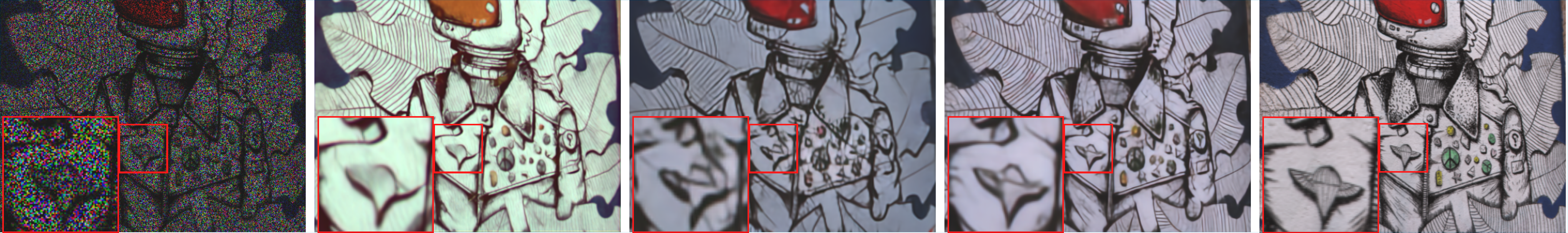}\\
    \vspace{0.1em}
    \includegraphics[width=\linewidth]{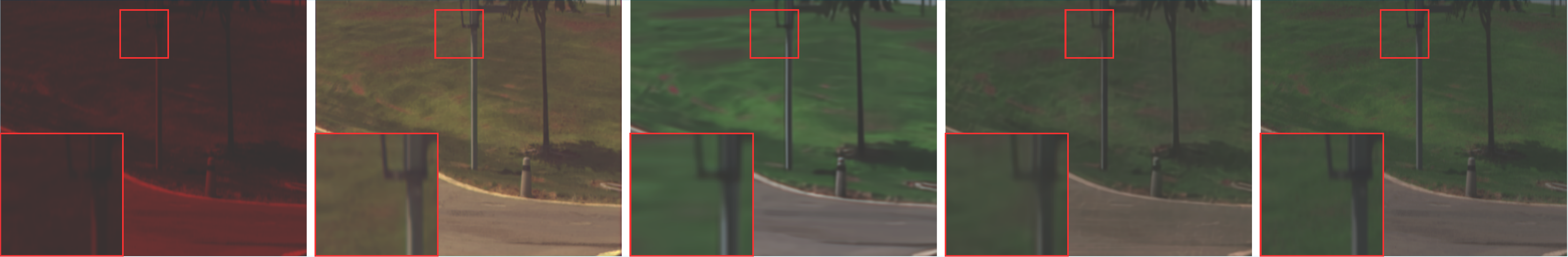}\\
    \vspace{0.1em}
    \includegraphics[width=\linewidth]{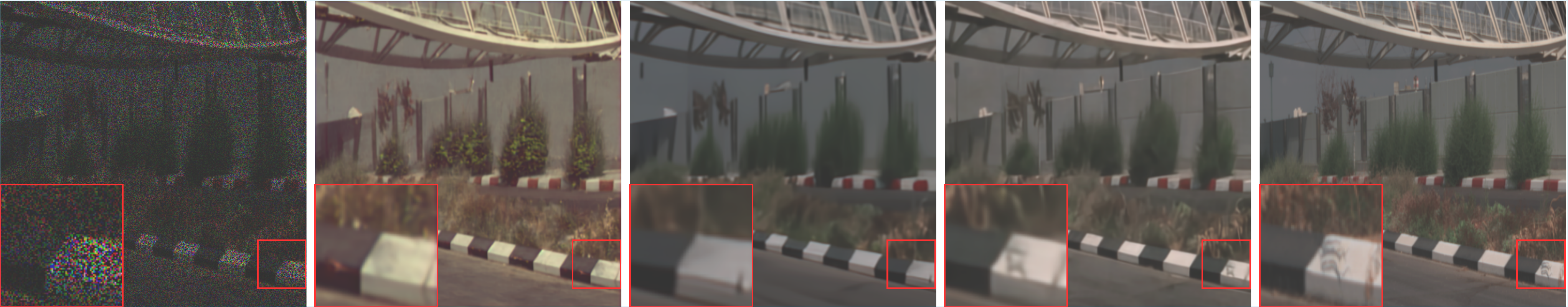}\\
    \noindent\makebox[1\linewidth]{
        \hspace{14mm}\textbf{LQ}\hfill
        \hspace{24mm}\textbf{PSRSCI}\hfill 
        \hspace{20mm}\textbf{VARSR}\hfill 
        \hspace{15mm}\textbf{HSI-VAR (Ours)}\hfill 
        \hspace{14mm}\textbf{\quad GT}
        \hspace{12mm}
    }
    \vspace{-4mm}
    \caption{Visual comparison with generative methods on different degradations.}
    \label{all_in_one_results}
    \vspace{-4mm}
\end{figure*}

\noindent \textbf{Metrics.}
We evaluate the performance by fidelity-based and perceptual quality measures: Peak Signal-to-Noise Ratio (PSNR), Structural Similarity Index Measure (SSIM), Learned Perceptual Image Patch Similarity (LPIPS), Deep Image Structure, Texture Similarity (DISTS) and three no-reference Image Quality Assessment (IQA) metrics (CLIP-IQA \cite{wang2022exploring}, MUSIQ \cite{ke2021musiq}, and MANIQA \cite{yang2022maniqa}).

\noindent \textbf{Implementation Details.}
We train our HSI-VAR using a GPT-2 style transformer with 16 blocks (pre-trained on class-to-image generation task) and a refiner with 6 blocks based on the NAFNet \cite{chen2022simple}.
All experiments are conducted on 8 NVIDIA GeForce RTX 4090 GPUs.
For VQVAE training, we adopt Adam optimizer with 32 batchsize for 300 epochs and the learning rate is $2e-5$.
For HSI-VAR, we leverage the AdamW optimizer with 32 batchsize for 150 epochs and the learning rate is $5e-5$.
Finetuning leverages 32 batchsize for 20 epochs with $2e-5$ learning rate.

\begin{table}[htbp]
\centering
\vspace{1mm}
\caption{Complexity comparisons on the generative models.}
\vspace{-2mm}
\label{tab:Complexity}
\resizebox{0.45\textwidth}{!}{
\begin{tabular}{l c c c c}
\toprule
\textbf{Method} & \textbf{Steps} & \textbf{TFLOPs} & \textbf{Params (M)} \\
\midrule
PSRSCI & 100 & 68.21 & 1312.00 \\
VARSR & 10 & 12.56 & 1210.96 \\
Ours (bare) & \textbf{10} & 2.67 & \textbf{466.82} \\
Ours (+DAG) & \textbf{10} & \textbf{1.38} & \textbf{466.82} \\
Ours (+DAG+Align)& \textbf{10} & 1.90 & 478.03 \\
Ours (+DAG+Align+SSA)& \textbf{10} & 1.91 & 483.00 \\
\bottomrule
\end{tabular}}
\vspace{-5mm}
\end{table}

\subsection{Experiment Results}
We compare our HSI-VAR with two regression methods (PromptIR \cite{potlapalli2306promptir} and VLUNet \cite{zeng2025vision}), diffusion-based method PSRSCI \cite{zeng2025spectral} and VAR-based method VARSR \cite{qu2025visual}.
We retrain or finetune all the compared methods on their official codes and checkpoints until convergence.

\begin{table*}
\centering
\caption{Ablation results of the 9-task all-in-one HSI restoration on \textbf{\textit{ICVL}} and \textbf{\textit{ARAD}} datasets.}
\vspace{-2mm}
\setlength{\tabcolsep}{2.5pt}
\label{tab:ablation}
\resizebox{\textwidth}{!}{ 
\begin{tabular}{ccccccccccccccccc}
    \toprule
    \multicolumn{3}{c}{\textbf{Module Configuration}} & \multicolumn{7}{c}{\textbf{\textit{ICVL}}} & \multicolumn{7}{c}{\textit{\textbf{ARAD}}} \\
    \cmidrule(lr){1-3} \cmidrule(lr){4-10} \cmidrule(lr){11-17} 
    \textbf{Alignment} & \textbf{DAG} & \textbf{SSA} & 
    \textbf{PSNR} $\uparrow$ & \textbf{SSIM} $\uparrow$ & \textbf{LPIPS} $\downarrow$ & \textbf{DISTS} $\downarrow$ & \textbf{CLIP-IQA} $\uparrow$ & \textbf{MUSIQ} $\uparrow$ & \textbf{MANIQA} $\uparrow$ & 
    \textbf{PSNR} $\uparrow$ & \textbf{SSIM} $\uparrow$ & \textbf{LPIPS} $\downarrow$ & \textbf{DISTS} $\downarrow$ & \textbf{CLIP-IQA} $\uparrow$ & \textbf{MUSIQ} $\uparrow$ & \textbf{MANIQA} $\uparrow$ \\ 
    \midrule
    $\times$ & $\times$ & $\times$ & \large 22.53 & \large .795 & \large .263 & \large .227 & \large .332 & \large 26.26 & \large .209 & \large 23.54 & \large .783 & \large .275 & \large .236 & \large .318 & \large 32.50 & \large .203 \\
    \midrule
    \checkmark & $\times$ & $\times$ & \large 32.28 & \large .907 & \large \textbf{.206} & \large .198 & \large .330 & \large 27.21 & \large .210 & \large 28.73 & \large .840 & \large .242 & \large .235 & \large .309 & \large 32.85 & \large .201 \\
    \checkmark & \checkmark & $\times$ & \large 32.91 & \large .910 & \large .210 & \large .195 & \large .351 & \large 27.70 & \large \textbf{.215} & \large 29.02 & \large .859 & \large .231 & \large \textbf{.215} & \large \textbf{.327} & \large 33.19 & \large .223 \\
    \checkmark & \checkmark & \checkmark & \large \textbf{33.23} & \large \textbf{.915} & \large .207 & \large \textbf{.193} & \large \textbf{.352} & \large \textbf{27.79} & \large .210 & \large \textbf{29.06} & \large \textbf{.862} & \large \textbf{.228} & \large \textbf{.215} & \large .326 & \large \textbf{33.22} & \large \textbf{.225} \\
    \bottomrule
\end{tabular}
}
\vspace{-4mm}
\end{table*}

\noindent \textbf{Comparison of HSI Restoration Results.}
In Table \ref{tab:comparison_non_generative} which presents the comparison with regression methods, our HSI-VAR demonstrates advanced performance on IQA metrics, such as $18.2\%$ improvement of CLIP-IQA on average compared with VLUNet.
This shows that our VAR-based HSI-VAR has better perceptual quality for real world restoration.
In Table \ref{tab:col_comparison} which presents the comparison with generative methods, our HSI-VAR achieves better results (for example we achieve 3.45 dB in PSNR and 0.072 in SSIM on \textbf{\textit{ICVL}}, we achieve 3.4 dB in PSNR and 1.6 in MUSIQ on \textbf{\textit{ARAD}}).
Visually, our results contain richer details compared with other generative methods (see Figure \ref{all_in_one_results} for details).
Basically, diffusion model PSRSCI often exhibits color shifts when confronting complex and severe degradations.
In contrast, our HSI-VAR can maintain more coordinated and adapted colors due to its scale-wise generation architecture and conditional alignment paradigm.

\begin{figure}
    \centering
    \hspace{-2mm}
    \includegraphics[width=0.95\linewidth]{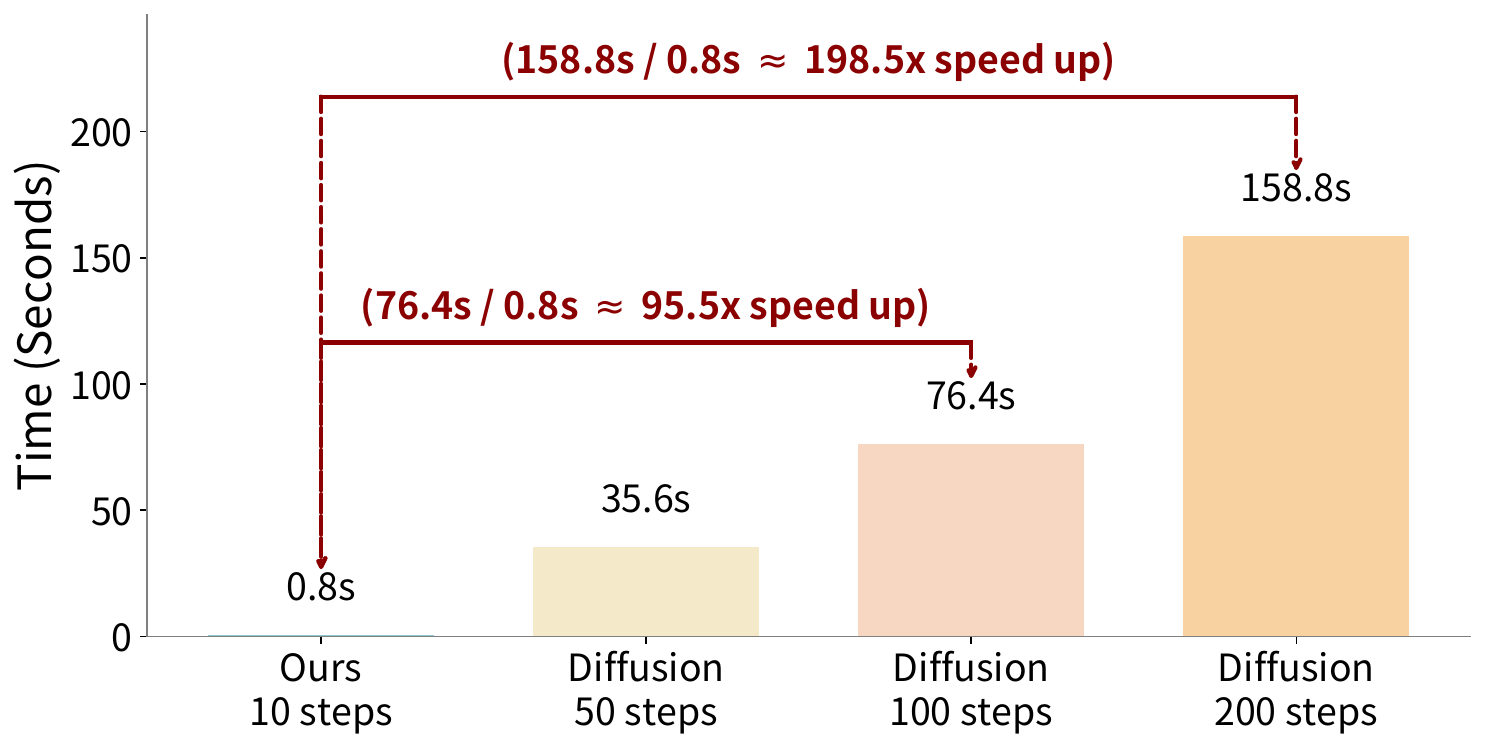}
    \vspace{-3mm}
    \caption{Comparisons on the inference time.}
    \vspace{-5mm}
    \label{fig:inference_time}
\end{figure}

\noindent \textbf{Complexity Analysis.}
VAR-based methods have the scale-wise generation strategy, which is much faster than diffusion-based methods which are affected by many sampling steps.
Table \ref{tab:Complexity} shows the comparison of the processing steps, Tera Floating-Point Operations Per Second (TFLOPs) and the parameters.
Our HSI-VAR requires not only half of the parameters than diffusion-based PSRSCI and VAR-based VARSR, but also nearly $10\%$ TFLOPs of VARSR.
For inference time in Figure \ref{fig:inference_time}, our VAR-based method achieves nearly $\times 95.5$ faster inference compared with the Diffusion-based PSRSCI with 100 steps.

\subsection{Ablation Analysis}
\label{sec:abalation}

\begin{figure}
    \centering
    \includegraphics[width=0.95\linewidth]{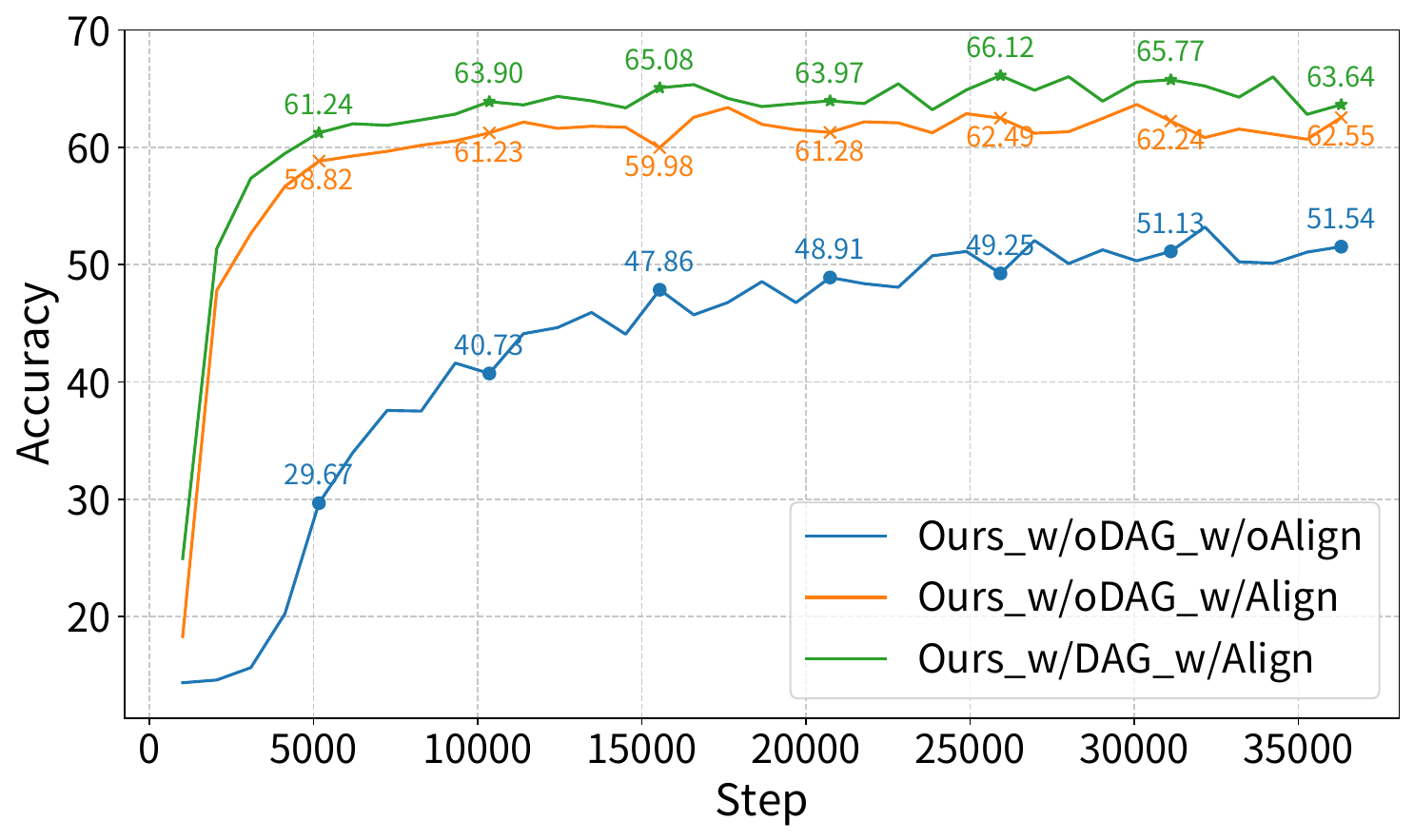}
    \vspace{-3mm}
    \caption{Comparisons on the accuracy in the validation.}
    \label{fig:accuracy}
    \vspace{-7mm}
\end{figure}
\noindent \textbf{Alignment Strategy Reinforces the Generation.}
In Table \ref{tab:ablation}, not only traditional fidelity-based metrics but also IQA metrics exhibit improvements (such as we improved PSNR by nearly 10 dB, LPIPS by 0.057 and MUSIQ by 0.95 on \textbf{\textit{ICVL}}) in the average results across all datasets and task settings with the proposed alignment strategy.
From Figure \ref{fig:accuracy}, we further observe that our HSI-VAR achieves faster convergence speed and a higher convergence peakupon.
These demonstrate that the proposed strategy bridges the gap between HQ inputs and degraded conditions and reinforces the overall generation process.

\noindent \textbf{DAG Models the Multi-degradation.}
In Table \ref{tab:ablation}, our HSI-VAR achieves 0.63 dB improvement in PSNR, 0.02 gain in CLIP-IQA, and 0.5 increase in MUSIQ with the proposed DAG. 
This demonstrates that our degradation modeling approach effectively handles diverse degraded features. 
In Figure \ref{fig:accuracy}, accuracy also improves with DAG, facilitating multi-scale generation. 
Table \ref{tab:Complexity} further shows that DAG cuts computational cost by $48.3\%$ requiring no extra parameters versus the regular CFG strategy, highlighting its efficiency.

\noindent \textbf{SSA Module Refines the Structure.}
Table \ref{tab:ablation} compares the performance with and without the proposed SSA module.
Notably, we observe improvements in metrics, indicating that SSA effectively refines and preserves structural details during the decoding phase, thereby facilitating visual autoregression from both spatial and spectral perspectives.

%% file: sec/5_conclusion.tex
\section{Conclusion}
We rethink HSI restoration as an autoregressive generation problem, where spectral and spatial dependencies can be progressively modeled rather than globally reconstructed.
Exising HSI restoration methods often suffer from hundreds of steps from the diffusion models and overly smooth results from the regression models.
Therefore we proposed the HSI-VAR, a novel VAR-based framework with the next scale prediction.
HSI-VAR proposes the following three strategies: latent–condition alignment strategy, DAG module and SSA module, which respectively bridge the embedding gap between LQ conditions and HQ latents, efficiently model the degradations with half-reduced computation and preserve the HSI structure.
Experiments demonstrate that our proposed HSI-VAR achieves superior performance on the all-in-one HSI restoration task.